\pdfoutput=1

\documentclass[11pt]{article}

\usepackage[]{ACL2023}

\usepackage{times}
\usepackage{latexsym}

\usepackage[T1]{fontenc}

\usepackage[utf8]{inputenc}

\usepackage{microtype}

\usepackage{inconsolata}

\usepackage{hyperref}
\usepackage{url}
\usepackage{booktabs}
\usepackage{multirow}
\usepackage{amssymb}
\usepackage{pifont}
\usepackage{tabularx}
\newcolumntype{Y}{>{\centering\arraybackslash}X}
\usepackage{adjustbox}
\usepackage{amsmath,amsfonts,bm}
\usepackage{lipsum}
\usepackage{comment}
\usepackage{algorithm2e}
\usepackage{subcaption}
\usepackage{diagbox}
\usepackage{cleveref}
\usepackage{longtable}
\usepackage{tikz}

\newcommand*\emptycirc[1][0.525ex]{\tikz\draw[thick] (0,0) circle (#1);} 

\DeclareMathOperator*{\argmax}{arg\,max}
\DeclareMathOperator*{\argmin}{arg\,min}

\newcommand{\gener}{GeNER}

\newcommand{\roberta}{RoBERTa}
\newcommand{\biobert}{BioBERT}
\newcommand{\roster}{RoSTER}
\newcommand{\bond}{\textsc{BOND}}
\newcommand{\quip}{\textsc{QuIP}}

\newcommand{\densephrases}{DensePhrases}
\newcommand{\typetoken}{[\texttt{TYPE}]}
\newcommand{\ours}{HighGEN}
\newcommand{\xone}{$\hat{\mathbf{X}}_1$}
\newcommand{\xtwo}{$\hat{\mathbf{X}}_2$}
\newcommand{\vone}{$\mathcal{\hat{V}}_1$}
\newcommand{\vtwo}{$\mathcal{\hat{V}}_2$}

\title{Automatic Creation of Named Entity Recognition Datasets by \\Querying Phrase Representations}

\author{Hyunjae Kim$^1$ \quad Jaehyo Yoo$^1$ \quad Seunghyun Yoon$^2$ \quad Jaewoo Kang$^{1,3}$\\
$^1$Korea University \quad $^2$Adobe Research \quad $^3$AIGEN Sciences \\
\texttt{\{hyunjae-kim,jaehyoyoo,kangj\}@korea.ac.kr} \\ \texttt{syoon@adobe.com}
}

\begin{document}
\maketitle

\begin{abstract}
Most weakly supervised named entity recognition (NER) models rely on domain-specific dictionaries provided by experts. This approach is infeasible in many domains where dictionaries do not exist. While a phrase retrieval model was used to construct pseudo-dictionaries with entities retrieved from Wikipedia automatically in a recent study, these dictionaries often have limited coverage because the retriever is likely to retrieve popular entities rather than rare ones. 
In this study, we present a novel framework, HighGEN, that generates NER datasets with high-coverage pseudo-dictionaries.
Specifically, we create entity-rich dictionaries with a novel search method, called phrase embedding search, which encourages the retriever to search a space densely populated with various entities. 
In addition, we use a new verification process based on the embedding distance between candidate entity mentions and entity types to reduce the false-positive noise in weak labels generated by high-coverage dictionaries. 
We demonstrate that HighGEN outperforms the previous best model by an average F1 score of 4.7 across five NER benchmark datasets.

\end{abstract}

\section{Introduction}
\label{sec:intro}

Named entity recognition (NER) models often require a vast number of manual annotations for training, which limits their utility in practice. In several studies, external resources such as domain-specific dictionaries have been employed as weak supervision to reduce annotation costs~\citep{shang-etal-2018-learning,liang2020bond,meng-etal-2021-distantly}. 
However, such dictionaries exist only for certain domains and building a dictionary for a new domain requires a high level of expertise and effort.

To address this problem, a recent study proposed a framework called \gener, which generates NER datasets without hand-crafted dictionaries~\cite{kim2021simple}.
In GeNER, user questions that reflect the needs for NER are received as inputs (e.g., ``\textit{Which city?}''), and an open-domain question-answering (QA) system, \densephrases~\cite{lee-etal-2021-learning-dense}, is used to retrieve relevant phrases (i.e., answers) and evidence sentences from Wikipedia. 
The retrieved phrases constitute a `pseudo' dictionary, which serves as weak supervision in place of hand-crafted dictionaries.
The evidence sentences are annotated based on string matching with the pseudo dictionary, resulting in the final dataset. 
This approach allows NER models to adapt to new domains for which training data are scarce and domain-specific dictionaries are unavailable.

\begin{figure}[t]

\begin{subfigure}{.99\columnwidth}
  \centering
  \includegraphics[width=.85\linewidth]{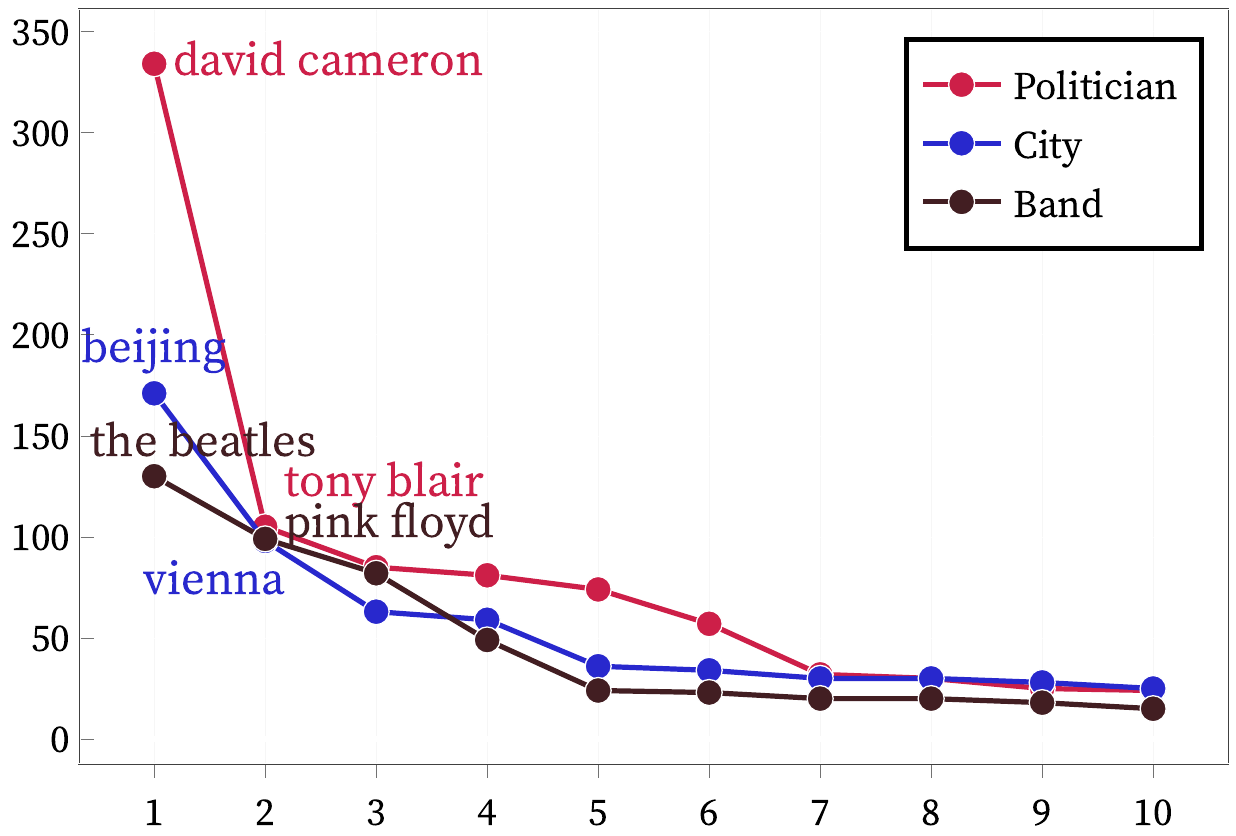} 
\end{subfigure}

\caption{
The most frequent ten entities in the top 1,000 phrases retrieved from the 2018-12-20 version of Wikipedia for the three questions: ``\textit{Which politician?}'', ``\textit{Which city?}'', and ``\textit{Which band?}''.
Due to a bias in the entity popularity~\cite{chen-etal-2021-evaluating}, a current phrase retrieval model, DensePhrases~\cite{lee-etal-2021-learning-dense}, primarily returns popular entities, limiting the coverage of dictionaries.
}
\label{fig:motivating}
\end{figure}

However, because the entity popularity of Wikipedia is biased~\cite{chen-etal-2021-evaluating,leszczynski-etal-2022-tabi}, existing open-domain QA models tend to retrieve popular entities rather than rare ones. This limits the coverage of dictionaries generated by GeNER. 
\Cref{fig:motivating}~shows examples of a bias in the entity population retrieved from the open-domain QA model. 
``David Cameron,'' ``Beijing,'' and ``The Beatles'' frequently appear in the top 1,000 retrieved phrases for each type of question.
Low-coverage dictionaries created from these biased results can cause incomplete annotations (i.e., false-negative entities), which impedes the training of NER models.
Unfortunately, increasing the number of retrieved phrases (i.e., larger top-k) is not an appropriate solution because it is computationally inefficient and causes a high false-positive rate in the dictionary. 
Therefore, a new search method that can efficiently retrieve diverse entities with a reasonable top-k and a new NER dataset generation framework based on this search method are needed.

In this study, we present \ours, an advanced framework for generating NER datasets with automatically constructed `high-coverage' dictionaries.
Specifically, we first obtain phrases and sentences and constitute an initial dictionary in a similar manner to GeNER. 
Subsequently, we expand the initial dictionary using a \textit{phrase embedding search}, in which the embeddings of the retrieved phrases are averaged to re-formulate query vectors.
These new queries specify contexts in which different entities of the same type appear, allowing our retriever to search over a vector space in which various entities are densely populated.\footnote{We provide an explanation of its working principle (\Cref{subsec:phrase_embedding_search}) and an analysis of its retrieval diversity (\Cref{subsec:retrieval_analysis}).}
The expanded dictionary is used to annotate the retrieved sentences. 
Because a larger dictionary can induce more false-positive annotations during rule-based string matching, we introduce a new verification process to ensure that weak labels annotated by the string matching are correct.
The verification process is performed by comparing the distance between the embeddings of a candidate entity and the target entity type.

We trained recent NER models~\cite{liu2019roberta,lee2020biobert,liang2020bond,meng-etal-2021-distantly} with the datasets generated by \ours~and evaluated the models on five datasets. 
Our models outperformed the baseline models trained using the previous best model \gener~by an average F1 score of 4.7 (\Cref{sec:experiments}).
In addition, we show an additional advantage of HighGEN over GeNER, which generates datasets using only a few hand-labeled examples without input user questions.
HighGEN outperformed few-shot NER models on two datasets (\Cref{sec:fewshot}).
Finally, we perform an analysis of the factors affecting the retrieval diversity and NER performance (\Cref{sec:analysis}).
We make the following contributions:\footnote{We will make our code publicly available.}
\begin{itemize}
    \item
    We propose a \ours~framework that generates NER datasets with entity-rich dictionaries that are automatically constructed from unlabeled Wikipedia corpus.
    \item 
    We present two novel methods in HighGEN:
    (i) phrase embedding search to overcome the limitations of the current open-domain phrase retriever and successfully increase the entity recall rate and (ii) distance-based verification to effectively reduce the false-positive noise in weak labels.
    \item 
    \ours~outperformed the previous-best weakly-supervised model GeNER by an F1 score of 4.7 on five datasets.
    In few-shot NER, \ours~created datasets using few-shot examples as queries and outperformed current few-shot NER models on two datasets. 
\end{itemize}

\section{Preliminaries}

\subsection{Weakly Supervised NER}
\label{subsec:weakly_supervised_ner}
The aim of NER is to identify named entities in text and classify them into predefined entity types. 
Let $\mathcal{D}=\{\mathbf{X},\mathbf{Y}\}$ be a dataset, where $\mathbf{X}=\{\mathbf{x}_n\}_{n=1}^{N}$ is a list of $N$ unlabeled sentences and $\mathbf{Y}=\{\mathbf{y}_n\}_{n=1}^{N}$ is a list of $N$ corresponding token-level label sequences.
While supervised learning relies on the human-annotated labels, $\mathbf{Y}$, to train models, in weakly supervised NER, the weak labels $\hat{\mathbf{Y}}$ are generated using string matching between a domain-specific dictionary, $\mathcal{V}$, built by experts and the unlabeled sentences, $\mathbf{X}$~\cite{yang-etal-2018-distantly,shang-etal-2018-learning,peng-etal-2019-distantly,cao-etal-2019-low,yang-katiyar-2020-simple,liang2020bond,meng-etal-2021-distantly}.
Hand-crafted labeling rules are utilized in another line of studies~\cite{fries2017swellshark,ratner2017snorkel,safranchik2020weakly,zhao-etal-2021-glara}; however, these rules are difficult to apply to new entity types.
Recently, \citet{kim2021simple}~proposed \gener, in which weak labels are generated with a pseudo-dictionary, $\hat{\mathcal{V}}$, created using a phrase retrieval model. We follow their approach but present an advanced framework for addressing the low-coverage problem and obtaining more entity-rich dictionaries and NER datasets.

\begin{figure*}[t]
\centering
 \includegraphics[width=.85\linewidth]{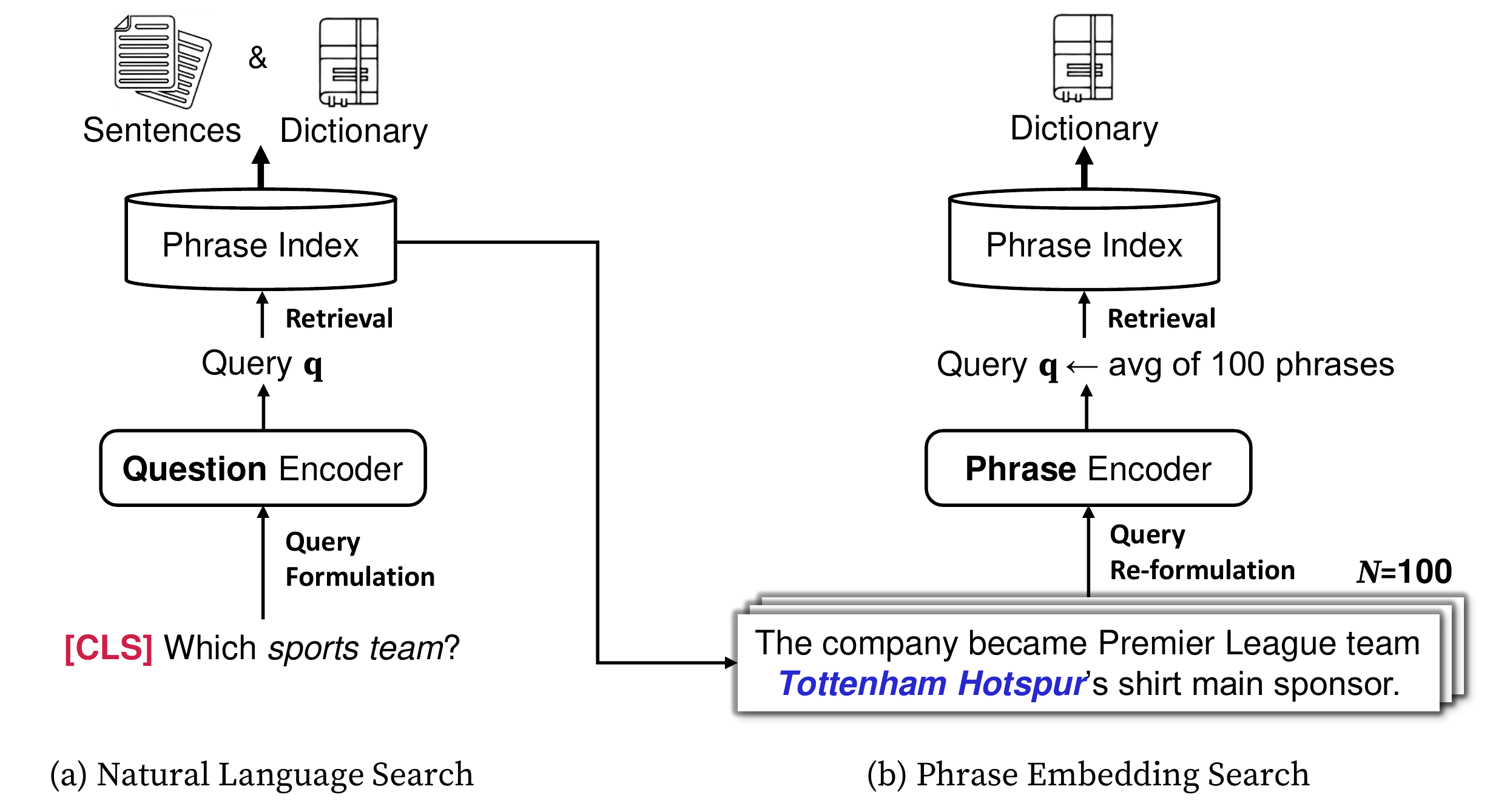}
\caption{
Overview of the natural language search and phrase embedding search in our \ours~framework. (a)~Natural language search retrieves unlabeled sentences ($\hat{\mathbf{X}}_\text{1}$) and an initial dictionary ($\hat{\mathcal{V}}_\text{1}$) for the given question, ``\textit{Which sports team?}'' (\Cref{subsec:natural_language_search}).
(b)~Phrase embedding search further retrieves an additional dictionary ($\hat{\mathcal{V}}_\text{2}$) using the top 100 phrases retrieved from the natural language search (\Cref{subsec:phrase_embedding_search}). 
Note that the retrieved sentences and phrases are fed into the dictionary matching and verification stage (see \Cref{subsec:embedding_matching} and \Cref{fig:verification}).
}
\label{fig:overview}
\end{figure*}

\subsection{DensePhrases}
\label{subsec:densephrases}

DensePhrases~\cite{lee-etal-2021-learning-dense} is a phrase retrieval model that finds relevant phrases for natural language inputs in a Wikipedia corpus.
Unlike the \textit{retriever-reader} approach, which first retrieves evidence passages from Wikipedia and then finds the answer~\cite{chen-etal-2017-reading}, DensePhrases retrieves answers directly from dense phrase vectors of the entire English Wikipedia as follows:
\begin{equation}
\label{equation:densephrases}
    \begin{aligned}
    &\mathbf{s} = E_{s}(s,x),\quad \mathbf{q} = E_{q}(q), \\
    &(s^*,x^*) = \argmax_{(s,x)\in\mathcal{W}}(\mathbf{s}^\top \mathbf{q}),
    \end{aligned}
\end{equation}
where $s$ is a phrase, a sequence of words from evidence text $x$ (i.e., sentence, passage, etc.);
$\mathcal{W}$ is the set of all phrase-evidence pairs in Wikipedia.
The input question $q$ is converted into the query vector $\mathbf{q}$ by the question encoder, $E_q$. 
Subsequently, relevant phrases are retrieved based on the similarity scores between the query vector $\mathbf{q}$ and phrase vector $\mathbf{s}$, which is represented as the concatenation of the start and end vectors of the phrase, produced by the phrase encoder, $E_s$. 
All phrase vectors are `pre-indexed' before inference, which greatly improves run-time efficiency~\cite{seo-etal-2019-real,lee-etal-2021-learning-dense}. 
In the context of weakly supervised NER, DensePhrases can be used as a database to obtain candidate entities for specific NER needs, along with sentences to construct the final NER corpus~\cite{kim2021simple}.

\subsection{Entity Popularity Bias}

\citet{chen-etal-2021-evaluating}~found that current document retrievers exhibit entity popularity bias in which the models prefer popular entities over rare ones and encounter problem in disambiguating entities in open-domain tasks. 
For instance, the models returned documents related to the company Apple for questions about the British rock band Apple or the 1980 film The Apple. 
Similarly, we raised the problem that phrase retrievers mainly provide popular entities for NER owing to the biased nature of Wikipedia in terms of entity popularity, which limits the coverage of dictionaries.

\section{Method}
\label{sec:highgen}

\begin{figure*}[t]
\centering
 \includegraphics[width=.975\linewidth]{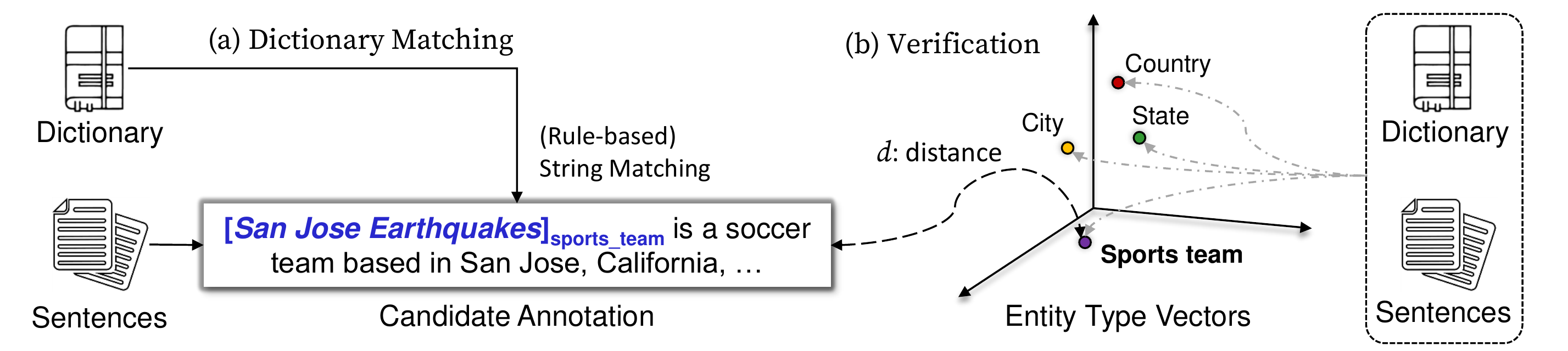}
\caption{
Overview of the (a) dictionary matching and (b) verification stage in our \ours~framework. 
This annotates sentences based on the vector distance between a candidate entity (e.g., ``San Jose Earthquakes'') and entity type (e.g, ``sport team'') that is represented as the average of phrase vectors in the retrieval results.
}
\label{fig:verification}
\end{figure*}

\ours~comprises three stages of natural language search, phrase embedding search (\Cref{fig:overview}), and dictionary matching and verification (\Cref{fig:verification}).
We highlight that the natural language search is similarly used in \gener, but the last two stages are novel and first proposed in our study.

\begin{table*}[t!]
\centering
\footnotesize
\begin{adjustbox}{max width = 0.99\textwidth}

\begin{tabular}{l}
\toprule
\textbf{Natural Language Search} \\
\midrule
\textbf{Q: Which actor?} \\
\relax[1] $\dots$ including Best British Film, Best British Director for Danny Boyle and Best British Actor for \textit{Ewan McGregor}. \\
\relax[2] His first movie role was in ``The Detective,'' which starred \textit{Frank Sinatra}. \\ 
\midrule
\textbf{Q: Which athlete?} \\
\relax[1] The nation's most famous Olympic athlete is \textit{Eric Moussambani}, who achieved some international notoriety for $\dots$ \\
\relax[2] \textit{Donovan Bailey} holds the men's world record with a time of 5.56 seconds and Irina Privalova holds the women's $\dots$ \\
\midrule
\textbf{Phrase Embedding Search} \\
\midrule
\textbf{Q: Which actor?} \\
\relax[1] \textit{Owen Ash Weingott} (21 June 1921 - 2013 12 October 2002) was an Australian actor and director although $\dots$, \\
\relax[2] \textit{Ron Vawter} (December 9, 1948 - 2013 April 16, 1994) was an American actor and a founding member of $\dots$, \\
\midrule
\textbf{Q: Which athlete?} \\
\relax[1] \textit{Yuri Floriani} (born 25 December 1981) is an Italian steeplechase runner. \\
\relax[2] \textit{Jeremy Porter Linn} (born January 6, 1975) is an American former competition swimmer, Olympic medalist, and $\dots$ \\
\bottomrule
\end{tabular}

\end{adjustbox}
\caption{
Comparison of context diversity of sentences retrieved using natural language search and phrase embedding search for the two questions.
Sentences by the phrase embedding search tend to have similar patterns.
}
\label{tab:context_analysis}
\end{table*}

\subsection{Natural Language Search}
\label{subsec:natural_language_search}

\paragraph{Query formulation.}
Let $T = \{t_1, ..., t_L\}$ be a set of $L$ target entity types.
The concrete needs for these entity types are translated into simple questions.
The questions follow the template of ``\textit{Which} $\typetoken$\textit{?},'' where the \typetoken~token is substituted for each entity type of interest. 
For instance, the question is formulated as ``\textit{Which} \texttt{city}\textit{?}'' if the target entity type $t$ is \texttt{city}.

\paragraph{Retrieval.}
Input questions are fed into the phrase retrieval model, DensePhrases, to retrieve the top $k$ phrases $s^{*}$ and sentences $x^{*}$ (see \Cref{subsec:densephrases}).
For $L$ different questions, a total of $k_1+\dots+k_L$ sentences are used as the unlabeled sentences, $\hat{\mathbf{X}}_\text{1}$.
The retrieved phrases are used as the pseudo-dictionary, $\hat{\mathcal{V}}_\text{1}$, which comprises phrase $s$ and corresponding type $t$ pairs (e.g., Beijing--city).

\subsection{Phrase Embedding Search}
\label{subsec:phrase_embedding_search}

\paragraph{Query re-formulation.}
As mentioned in \Cref{sec:intro}, the coverage of the initial dictionary $\hat{\mathcal{V}}_\text{1}$ is often limited because of the entity popularity bias.
Our solution to search for diverse entities is very simple.
We re-formulate queries by averaging the phrase vectors as follows:
\begin{equation}
\label{equation:phrase_embedding_search}
    \mathbf{q} = \frac{1}{N}\sum^{N}_{n=1} E_s(s_n, x_n),
\end{equation}
where $s_n$ and $x_n$ are the $n$-th top phrase and corresponding sentence from the natural language search.
We used only the top 100 phrases for each question (i.e., $N=100$) because a larger number of phrases did not improve retrieval quality in our initial experiments. 

\paragraph{Retrieval.}
For $L$ new queries obtained by \Cref{equation:phrase_embedding_search}, a total of $k^{\prime}_1+\dots+k^{\prime}_L$ phrases are additionally retrieved by \Cref{equation:densephrases} and constitute a new dictionary $\hat{\mathcal{V}}_\text{2}$. 
Subsequently, we merge $\hat{\mathcal{V}}_\text{1}$ and $\hat{\mathcal{V}}_\text{2}$ to obtain the final dictionary $\hat{\mathcal{V}}$. 
Note that we do not use the retrieved sentences $\hat{\mathbf{X}}_\text{2}$ because we found using only $\hat{\mathbf{X}}_\text{1}$ as the final unlabeled sentences (i.e., $\hat{\mathbf{X}}$) resulted in better NER performance.\footnote{A related analysis is included in \Cref{subsec:data_size}.}

\paragraph{Interpretation.}


Natural language search results in the retriever performing `broad' searches for all the Wikipedia contexts relevant to the target entity class. 
In contrast, phrase embedding search, which averages phrase vectors of the same entity type, can be viewed as providing prompts that implicitly represent certain contextual patterns in which entities of the target class often appear. 
Having the retriever perform `narrow' searches by focusing on specific contexts leads to a wide variety of entities with less bias towards popular ones.
This is because (1) the same entities rarely appear repeatedly in a specific context, (2) whereas different entities of the same type frequently appear in a similar context as they are generally interchangeable.

Our qualitative analysis supports our claim above. 
We retrieved 5k sentences using two questions, ``\textit{Which actor?}'' and ``\textit{Which athlete?}'', and manually analyzed 100 sentences sampled from them. 
\Cref{tab:context_analysis} shows that sentences by the phrase embedding search exhibit clear patterns in their contexts, whereas those by the natural language search do not.
Specifically, 91 and 94 of the 100 sentences for the actor and athlete types had similar patterns, respectively.
Further analysis shows that this property of the phrase embedding search contributes significantly to improving entity diversity (\Cref{subsec:retrieval_analysis}) and NER performance (\Cref{subsec:data_size}).

\subsection{Dictionary Matching \& Verification}
\label{subsec:embedding_matching}

\paragraph{Dictionary matching.}
After $\hat{\mathbf{X}}$ and $\hat{\mathcal{V}}$ are obtained, dictionary matching is performed to generate weak labels, $\hat{\mathbf{Y}}$. 
Specifically, if a string in the unlabeled sentence matches an entity name in the dictionary, the string is labeled with the corresponding entity type.
However, this method cannot handle \textit{label ambiguity} inherent in entities\footnote{Even the same string can be labeled with different entity types depending on the context, leading to label ambiguity. For instance, ``Liverpool'' could be a city or a sports team.} because it relies only on lexical information without leveraging contextual information of phrases. 
The false-positive noise due to label ambiguity is amplified as the dictionary size increases, making it difficult to effectively use our expanded dictionary $\hat{\mathcal{V}}$.

\paragraph{Verification.}

Candidate annotations provided by dictionary matching are passed to the verification stage.
Let $e$ be a matched string in the sentence and $\bar{T}$ be the matched entity types (a subset of $T$).
The verification function $\mathcal{L}$ is defined as follows:
\begin{align}
\label{equation:verification}
    & \mathcal{L}(e,\bar{T}) = 
    \begin{cases}
        t^* & \text{if } d(\mathbf{e},\mathbf{t}^*) < \tau, \\
        \text{not entity}        & \text{otherwise,}
    \end{cases} \\
    & t^* = \argmin_{t_l \in \bar{T}} d(\mathbf{e},\mathbf{t}_l), \; \mathbf{t}_l = \frac{1}{k_l} \sum^{k_l}_{n=1} E_s(s_n,x_n),\notag
\end{align}
where $d$ is the Euclidean distance function; $\mathbf{e}$ is the phrase vector of the candidate string; $\mathbf{t}_l$ is the $l$-th type vector; $\tau$ is the cut-off value.
The string is labeled with the nearest type $t^*$, or unlabeled if the distance is higher than the cut-off value.
The type vector is calculated by averaging all the retrieved phrase vectors of the entity type, based on the assumption that the mean vector of phrases is a good representative of the entity class.
In addition, the cut-off value is also calculated using phrase vectors. 
Specifically, the function $d$ computes the distance scores between the type vector $t_l$ and all the phrase vectors of the type. 
The distribution of the distance scores is then standardized, and the score of `$z$' times the standard deviation from the mean is used as the cut-off value (e.g., $z=3$).\footnote{The distribution of the distance scores is generally balanced; thus, we used a usual method to compute the cut-off value without any other tricks to balance the distribution.}

\section{Experiments}
\label{sec:experiments}

In this experiment, it was assumed that human-annotated datasets did not exist; thus, our models were trained only using synthetic data \{$\hat{\mathbf{X}}, \hat{\mathbf{Y}}$\} by \ours. 
To avoid excessive hyperparameter search, we used the same sets of input questions and the same number of sentences for each question (i.e., $k_1, \dots, k_L$) as those used in the previous study~\cite{kim2021simple}.
A new hyperparameter introduced in \ours, the number of phrases retrieved by phrase embedding search (i.e., $k^{\prime}_1,\dots,k^{\prime}_L$), was set to 30k.
Please refer to in \Cref{appendix:setups}~for the full list of hyperparameters and implementation details.
For metrics, the entity-level precision, recall, and F1 scores were used~\cite{tjong-kim-sang-de-meulder-2003-introduction}. 

\subsection{Datasets}
We used five datasets from four domains.
Following \citet{kim2021simple}, we did not use the \textit{MISC} and \textit{other} classes because they are vague to represent with some user questions.
(i) {CoNLL-2003}~\citep{tjong-kim-sang-de-meulder-2003-introduction} consists of Reuters news articles with three coarse-grained entity types of person, location, and organization.
(ii) {Wikigold}~\citep{balasuriya-etal-2009-named} is a small-size dataset that consists of Wikipedia documents with the same entity types as CoNLL-2003.
(iii) {WNUT-16}~\cite{strauss-etal-2016-results} consists of nine entity types annotated in tweets, such as TV show, movie, and musician.
(iv) Two biomedical domain datasets, NCBI-disease~\citep{dougan2014ncbi} and BC5CDR~\citep{li2016biocreative}, are collections of PubMed abstracts with manually annotated diseases (NCBI-disease) or disease and chemical entities (BC5CDR).
The benchmark statistics are listed in \Cref{tab:statistics} (Appendix).

\begin{table*}[t!]
\centering
\footnotesize

\begin{tabular}{lccccc}
\toprule
\multirow{1}{*}{\textbf{Model}} & \multicolumn{1}{c}{\textbf{CoNLL-2003}} & \multicolumn{1}{c}{\textbf{Wikigold}} & \multicolumn{1}{c}{\textbf{WNUT-16}} & \multicolumn{1}{c}{\textbf{NCBI-disease}} & \multicolumn{1}{c}{\textbf{BC5CDR}} \\
\midrule
\multicolumn{6}{l}{Full Dictionary} \\
\quad+ Standard & 74.4 (80.5/69.1) & 54.9 (53.8/56.1) & 45.3 (44.3/46.2) & 66.6 (67.5/65.7) & 79.7 (82.8/76.8)  \\
\quad + \bond & 83.5 (82.1/84.9) & 55.7 (46.0/70.8) & 35.0 (30.6/40.9) & 67.0 (63.7/70.6) & 81.1 (76.6/86.1)  \\
\quad + \roster & 85.8 (84.3/87.3) & 73.1 (67.1/80.2) & 28.9 (43.1/21.8) & 74.3 (75.9/72.7) & 80.7 (78.6/83.0)  \\
\midrule
\multicolumn{6}{l}{{Pseudo-dictionary}} \\
\midrule
\multicolumn{6}{l}{GeNER} \\
\quad + Standard & 56.3 (72.9/45.8) & 41.3 (58.6/31.9) & 36.5 (41.3/32.6) & 45.9 (59.0/37.6) & 64.9 (76.6/56.3)  \\
\quad + \bond & 64.5 (70.7/59.3) & 59.5 (65.2/54.7) & 42.1 (36.7/49.5) & 67.0 (70.8/63.5) & 69.3 (69.0/69.7)  \\
\quad + \roster & 67.8 (77.9/60.0)	& 55.8 (66.9/47.9) & 51.8 (49.1/54.8) & 71.0 (74.1/68.1) & 72.1 (74.6/69.7)  \\
\midrule
\multicolumn{6}{l}{\ours~(\textbf{Ours})} \\
\quad+ Standard & 58.0 (73.3/48.0) & 43.6 (59.5/34.4) & 38.5 (42.2/35.4) & 53.3 (66.4/44.6) & 72.2 (77.9/67.3)  \\
\quad+ \bond & 66.0 (65.5/66.5) & \textbf{68.2} (67.2/69.2) & 40.2 (32.6/52.3) & 70.2 (72.9/67.6) & 72.9 (69.5/76.7) \\
\quad+ \roster & \textbf{73.3} (78.5/68.7) & 67.5 (68.5/66.5) & \textbf{53.4} (49.0/58.8) & \textbf{73.2} (77.4/69.4) & \textbf{74.6} (73.3/76.0) \\
\midrule
\multicolumn{6}{l}{\ours~+ \roster~(for ablation study)} \\
\qquad w/o $\mathcal{L}$ & 70.6 (68.2/73.1) & 65.7 (56.8/78.0) & 35.1 (24.3/63.6) & 71.4 (69.7/73.2) & 72.2 (68.3/76.6) \\
\bottomrule
\end{tabular}
\caption{
Main results on the test sets of five NER benchmarks.
F1 score (precision/recall) is reported.
}
\label{tab:result_main}

\end{table*}

\subsection{NER Models}
\label{subsec:ner_models}
We trained three types of NER models on our synthetic data.
We provide descriptions of the models below, but we cannot cover all the details; readers interested in details are therefore recommended to refer to~\citet{liang2020bond}~and~\citet{meng-etal-2021-distantly}.
Note that we did not use validation sets to find the best model parameters during training to avoid excessive parameter tuning.
The implementation details are provided in \Cref{appendix:setups}.

\paragraph{Standard\textmd{:}} This type of model consists of a pre-trained language model for encoding input sequences and a linear layer for token-level prediction. 
We used RoBERTa~\cite{liu2019roberta} as the language model for the news, Wikipedia, and Twitter domains and BioBERT~\cite{lee2020biobert} for the biomedical domain.

\paragraph{\bond~\textmd{\cite{liang2020bond}:}}
This model is based on {self-training}, which is a learning algorithm that corrects weak labels with the power of large-scale language models.
Specifically, a \textit{teacher} model (similar to the standard model above) is initially trained on the weakly-labeled corpus and used to re-annotate the corpus based on its predictions.
This re-annotation process allows the model to remove noisy labels and further identify missing entities.
A \textit{student} model with the same model structure as the teacher model is trained on the re-annotated corpus.
The teacher model is updated by the student model's parameters in the next round and performs the re-annotation process again.
This process is repeated until the maximum training step is reached.

\paragraph{\roster~\textmd{\cite{meng-etal-2021-distantly}:}} 

In \roster, the \textit{generalized cross-entropy (GCE)} loss is used to a standard model, which is designed to be more robust to noise than a normal cross-entropy loss. 
During the GCE training, weak labels are removed at every update step if the model assigns low confidence scores.
Using the algorithm described above, five randomly initialized models are trained, and a new model is trained to approximate the average predictions of the five models. 
Finally, the new model is further trained with \textit{language model augmented self-training}, which jointly approximates the teacher model's predictions for the given (1) original sequence and (2) augmented sequence with some tokens replaced by a language model.

\subsection{In-domain Resources}
Baseline models are classified into two categories based on the amount of in-domain resources required during training.

\paragraph{\gener~\textmd{\cite{kim2021simple}:}} GeNER is the only baseline model that uses the same amount of resources as \ours.
GeNER retrieves phrases and unlabeled sentences using natural language search and performs string matching to create datasets.

\paragraph{Full dictionary\textmd{:}} Full-dictionary models use large-scale dictionaries that comprises numerous entities hand-labeled by experts.
For the CoNLL-2003, Wikigold, and WNUT-16 datasets, each dictionary was constructed using Wikidata and dozens of gazetteers compiled from multiple websites~\cite{liang2020bond}.
For NCBI-disease and BC5CDR, the dictionary was constructed by combining the MeSH database and Comparative Toxicogenomics Database (more than 300k disease and chemical entities)~\cite{shang-etal-2018-learning}.
These dictionaries were used to generate weak labels based on string matches with \textit{in-domain corpus}, which is an unlabeled version of the original training corpus.

\subsection{Results}
\Cref{tab:result_main}~shows that HighGEN outperformed GeNER on five datasets by average F1 scores of 4.2, 3.0, and 4.7 for the standard, \bond, and \roster~models, respectively.
Performance improvements were particularly evident in recall.
When the verification method was not applied (i.e., w/o $\mathcal{L}$), the performance dropped by an average F1 score of 5.4 (mostly in precision).
A high NER performance can be expected with full dictionaries, but they cannot be built without tremendous effort of experts. 
We emphasize that our method of automatically creating high-coverage pseudo-dictionaries and NER datasets is a promising way to achieve competitive performance with minimal effort.

\section{Few-shot NER}
\label{sec:fewshot}

We show an additional use case for HighGEN to create NER datasets using only a few hand-labeled examples, without using input questions.
This can eliminate a tuning/engineering effort of users that might be required for designing appropriate questions to identify NER needs, which is a distinct advantage of \ours~over GeNER.
Specifically, \ours~takes sentences with annotated phrases as input and retrieves \xtwo~and \vtwo~using the phrase embedding search (defined in Equations (\ref{equation:densephrases}) and (\ref{equation:phrase_embedding_search})), which are used as the unlabeled sentences and pseudo-dictionary to produce the final dataset.

We tested two types of models. 
(1) The entity-level model uses every annotated phrase as a separate query; thus, the number of queries equals the number of human annotations.
On the other hand, (2) the class-level model first averages phrase vectors of the same entity types and uses them as queries; thus, the number of queries equals the number of entity types.
The entity-level model would have an advantage in terms of entity recall and the class-level model can mitigate noise that each phrase vector may contain.

\begin{table}[t!]
\centering
\footnotesize
\begin{adjustbox}{max width = 0.99\columnwidth}

\begin{tabular}{lcc}
\toprule
\textbf{Model} & \textbf{CoNLL-2003} & \textbf{BC5CDR} \\
\midrule
\multicolumn{3}{l}{{5-shot sentences (per entity type)}} \\
\midrule
Supervised & 53.5 & 55.0 \\
\quad + NSP & 61.4 & - \\
\qquad + Self-training & 65.4 & - \\
\quip~& 74.0 & 65.7 \\
\midrule
HighGEN (entity) & \textbf{75.6} & 68.2 \\
HighGEN (class) & 73.2 & \textbf{72.5} \\
\bottomrule
\end{tabular}
\end{adjustbox}
\caption{
F1 scores of \ours~and baseline models in few-shot NER.
Note that the scores of the baseline models are from previous studies~\cite{huang-etal-2021-shot,jia-etal-2022-question,kim2021simple}.
}
\label{tab:fewshot_results}
\end{table}

\paragraph{Setups.}
We sampled datasets from CoNLL-2003 and BC5CDR so that each dataset consists of five sentences per entity type, which results in 20 and 10 examples for CoNLL-2003 and BC5CDR, respectively.
\footnote{Unlike the experiments in \Cref{sec:experiments}, the \textit{MISC} type was included for a fair comparison with baseline models.}
All experimental results were averaged over five sampled datasets.
We used the models of \citet{huang-etal-2021-shot} and \citet{jia-etal-2022-question} as baselines, and among them, \quip~\cite{jia-etal-2022-question} is the previous best model in few-shot NER
(details on the models are presented in \Cref{appendix:fewshot}).\footnote{Other few-shot NER models were excluded because they used a sufficient amount of `source' data~\cite{yang-katiyar-2020-simple,cui-etal-2021-template}, which is different from our setups.}
For \ours, we retrieved the same number of sentences for each query, and the total number of sentences was 120k for CoNLL-2003 and and 10k for BC5CDR.
We initially trained \roster~on our synthetic data and then fine-tuned the model on few-shot examples.

\paragraph{Results.}
\Cref{tab:fewshot_results} shows that our entity- and class-level models outperformed \quip~by an average F1 score of 2.1 and 3.0 on the two datasets, respectively.
For CoNLL-2003, the entity-level model was better than the class-level model because entities of the same entity type often belong to different sub-categories.
For instance, ``Volkswagen'' and ``University of Cambridge'' belong to the same \textit{organization} type in CoNLL-2003 but their sub-categories are ``company'' and ``institution,'' respectively.
Therefore, it is difficult to group them into a single vector and it is important to widely cover various entities using separate queries for each sub-category.
On the other hand, entities in BC5CDR can be naturally grouped by disease or chemical type, which allows the class-level model to perform well.
Additionally, biomedical entity names often contain domain-specific terms, numbers, special characters, and abbreviations that are difficult to encode with a general-purpose phrase encoder, making their vector representations relatively more error-prone.
The class-model can produce good representations by averaging phrase vectors.

\section{Analysis}
\label{sec:analysis}

\subsection{Retrieval Performance}
\label{subsec:retrieval_analysis}

We compared natural language search and phrase embedding search in terms of their accuracy and diversity. 
With reference to \citet{kim2021simple}, we used 11 fine-grained questions within the following four coarse-grained entity types of (i) person (athlete, politician, actor), (ii) location (country, city, state in the USA), (iii) organization (sports team, company, institution), and (iv) biomedicine (disease, drug). 
We report the average scores for each coarse-grained entity type.

\begin{table}[t!]
\centering
\footnotesize

\begin{adjustbox}{max width = 0.99\columnwidth}

\begin{tabular}{lccrccr}
\toprule
\multirow{3}{*}{\textbf{Type}} & \multicolumn{3}{c}{\textbf{P@100}} &  \multicolumn{3}{c}{\textbf{Div@10k}} \\
\cmidrule(lr){2-4} \cmidrule(lr){5-7}
 & NL & PE & \multicolumn{1}{c}{\enspace$\triangle$} & NL & PE & \multicolumn{1}{c}{$\triangle$} \\
\midrule
PER & 98.3 & 99.0 & 0.7 & 37.1 & 81.4 & 44.3  \\
LOC & 97.7 & 99.0 & 1.3 & \enspace2.6 & \enspace2.9 & 0.3 \\
ORG & 95.7 & 95.0 & $-$0.7 & 36.2 & 61.5 & 25.3 \\
BIO & 96.5 & 98.5 & 2.0 & 20.1 & 62.7 & 42.6 \\
\midrule
Average & 97.1 & 97.9 & 0.8 & 24.0 & 52.1 & 28.1 \\
\bottomrule
\end{tabular}
\end{adjustbox}
\caption{
Retrieval results from natural language search (NL) and phrase embedding search (PE) for the four entity types: person (PER), location (LOC), organization (ORG), and biomedicine (BIO).
$\triangle$: performance difference between PE and NL.
}
\label{tab:diversity}
\end{table}

\paragraph{Metrics.}
(i) The precision at 100 (P@100) represents the accuracy of the top 100 retrieved phrases. 
Because there are no gold annotations for the retrieved phrases, we manually determined whether the phrases correspond to the correct entity types.
(ii) Diversity at 10k (Div@10k) calculates the percentage of unique phrases out of the top 10k phrases based on their lowercase strings.

\paragraph{Results.}
The phrase embedding search largely outperformed the natural language search by a macro average of 28.1 diversity across the four types without loss of accuracy.
The diversity scores for the location entity types did not improve significantly because there are only limited numbers of names for locations such as countries in the real world, but the diversity scores for the other types improved dramatically (+ 37.4 diversity).

\begin{figure}[t]
\centering

\begin{subfigure}{.47\textwidth}
  \centering
  \includegraphics[width=.975\linewidth]{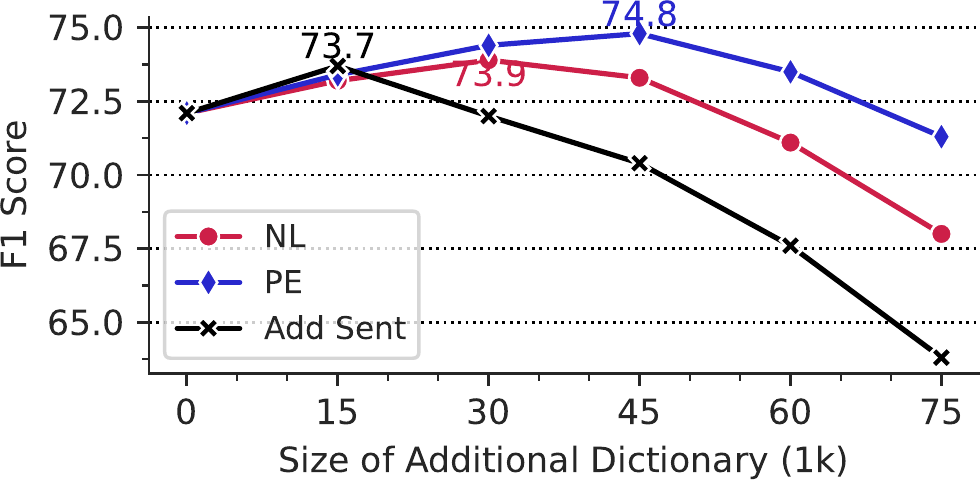}
  \label{subfig:dict_size}
\end{subfigure}

\begin{subfigure}{.47\textwidth}
  \centering
  \includegraphics[width=.95\linewidth]{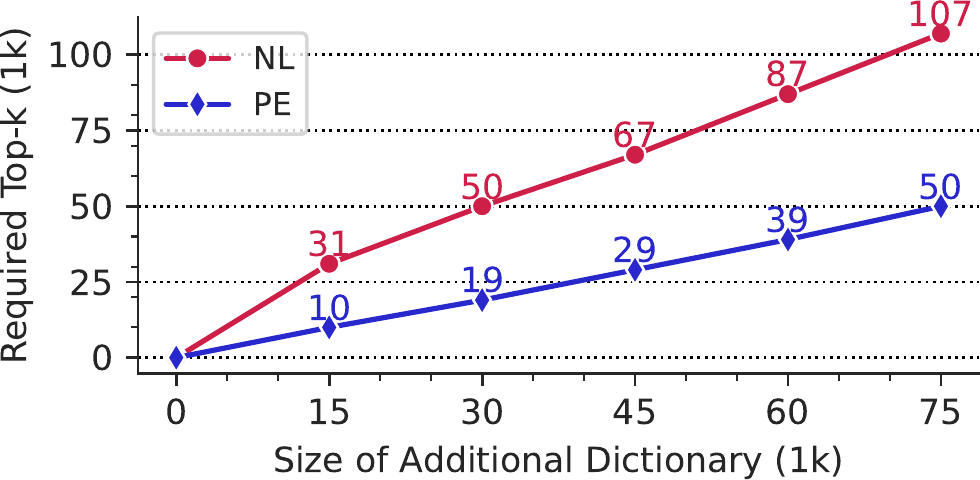} 
  \label{subfig:top_k}
\end{subfigure}

\caption{
Performance of \roster~models on BC5CDR with different sizes of the additional dictionary and the top-k to reach a certain dictionary size by the natural language search (NL) and phrase embedding search (PE).
`Add Sent' presents the performance of the model trained with additional sentences by the phrase embedding search (i.e., \xone$+$\xtwo).
The size of the initial dictionary (${x=0}$) is 12k.
}
\label{fig:dict_size}
\end{figure}

While both queries produced accurate top results (P@100), the accuracy tends to decrease as the top-k increased, which makes it difficult to increase the dictionary size by retrieving more phrases. 
Thus, retrieving diverse entities with a reasonable top-k is not only important for computational efficiency but also helps the retriever to maintain accuracy.
In this regard, phrase embedding search has a huge advantage over natural language search.
We discuss this further in \Cref{subsec:data_size}.
In addition, examples of the top phrases retrieved by both search methods are listed in \Cref{tab:retrieved_entities} (Appendix).

\begin{table*}[t!]
\centering
\footnotesize

\begin{tabular}{llll}
\toprule
\textbf{Sentence} & \begin{tabular}[c]{@{}c@{}}\textbf{Small \boldsymbol{$\mathcal{V}$}} \\ \textbf{+ String} \end{tabular} & \begin{tabular}[c]{@{}c@{}}\textbf{Large \boldsymbol{$\mathcal{V}$}} \\ \textbf{+ String} \end{tabular} & \begin{tabular}[c]{@{}c@{}}\textbf{Large \boldsymbol{$\mathcal{V}$}} \\ \textbf{+ Verif.} \end{tabular} \\
\midrule
\relax[1] $\dots$ and on \textbf{\underline{Central}} and Eastern Europeans living in the $\dots$  & None (\textcolor[HTML]{2828CD}{$\emptycirc$}) & Company (\textcolor[HTML]{CD1F48}{×}) & None (\textcolor[HTML]{2828CD}{$\emptycirc$})  \\
\relax[2] $\dots$ Foreign Minister \textbf{\underline{Alexander Downer}} and various $\dots$ & None (\textcolor[HTML]{CD1F48}{×}) & Politician (\textcolor[HTML]{2828CD}{$\emptycirc$}) & Politician (\textcolor[HTML]{2828CD}{$\emptycirc$})  \\
\relax[3] $\dots$ at club level for Cruzeiro, PSV, \textbf{\underline{Barcelona}}, and Inter Milan. & Sport team (\textcolor[HTML]{2828CD}{$\emptycirc$}) & City (\textcolor[HTML]{CD1F48}{×}) & Sport team (\textcolor[HTML]{2828CD}{$\emptycirc$}) \\
\bottomrule
\end{tabular}

\caption{
Case study of dictionary sizes and dictionary matching methods.
{Small $\mathcal{V}$}: initial dictionary (i.e., $\hat{\mathcal{V}}_\text{1}$) consisting of 12k entities.
{Large $\mathcal{V}$}: expanded dictionary (i.e., $\hat{\mathcal{V}}_\text{1}$ + $\hat{\mathcal{V}}_\text{2}$) consisting of 134k entities.
{String}: rule-based string matching.
{Verif.}: the verification method.
\textcolor[HTML]{CD1F48}{{×}}: incorrect annotations.
\textcolor[HTML]{2828CD}{$\emptycirc$}: correct annotations. 
}
\label{tab:case_study}
\end{table*}

\subsection{Data Size}
\label{subsec:data_size}

\paragraph{Effect of dictionary size.}
\Cref{fig:dict_size}~shows the NER performance of \roster~models according to the size of the additional dictionary added to the initial dictionary \vone. 
We expanded the dictionary using the natural language search or phrase embedding search.
F1 scores were measured on the BC5CDR test set.

The performance of both models increased initially but decreased after the peaks, indicating that there was a trade-off between the size and accuracy of the dictionary.
The optimal size of the additional dictionary by the phrase embedding search (i.e., 45k) was larger than that of the natural language search (i.e., 30k).
As shown in the second graph in \Cref{fig:dict_size}, the natural language search required a much larger number of sentences (more than twice as much) than the phrase embedding search to obtain the required dictionary size, which caused more false-positive results to be included in the dictionary.

\paragraph{Effect of Additional Sentences.}

In addition to using the additional dictionary \vtwo~obtained using phrase embedding search, we tried to use additional sentences \xtwo~along with \xone~(see `Add Sent' in \Cref{fig:dict_size}).
The performance was higher than the other models at low top-k ($x = 15$k), but the performance degraded rapidly as the dictionary size grew. 
As discussed in \Cref{subsec:phrase_embedding_search}, the sentences from the phrase embedding search have similar patterns, and from this result, we conjecture that the limited contextual patterns hindered the model's generalizability.
In conclusion, using only \xone~for the unlabeled corpus and both \vone~and \vtwo~for the dictionary would result in the best NER performance in most cases.
However, as shown in \Cref{sec:fewshot}, using \xtwo~and \vtwo~can be a good alternative if users want to avoid effort required in query tuning.

\subsection{Case Study}
\label{subsec:case_study}
\Cref{tab:case_study} shows several examples of how a large dictionary induced noise annotations in dictionary matching and how these annotations were corrected by the verification method. 
We used nine fine-grained entity types belonging to the person, location, and organization types, which were used in the experiments in \Cref{subsec:retrieval_analysis}.
We denote the initial dictionary (i.e., $\hat{\mathcal{V}}_\text{1}$) as a small dictionary and the expanded dictionary that consists of the initial and additional dictionaries (i.e., $\hat{\mathcal{V}}_\text{1}$ + $\hat{\mathcal{V}}_\text{2}$) as a large dictionary.
While the small dictionary could not match the entity ``Alexander Downer'' owing to its limited coverage, the entity was correctly annotated by a large dictionary. However, the large dictionary incorrectly annotated ``Central'' as a company, indicating that there is a trade-off between the coverage and accuracy of a dictionary. 
Also, ``Barcelona'' appeared mainly as a sports team in the small dictionary, whereas in the large dictionary it frequently appeared as a city and was therefore incorrectly annotated by the latter. 
In contrast, our verification method had the advantages of both dictionaries; it preserved the high accuracy of the small dictionary while retaining the high coverage of the large dictionary, resulting in correct annotations.

\section{Conclusion}

In this study, we presented an advanced dataset generation framework, HighGEN, which combines (1) phrase embedding search to address the problem of efficiently retrieving various entities using an open-domain retriever and (2) verification method to deal with false positives in a large dictionary.
In the experiments, we demonstrated the superiority of \ours~using five NER benchmarks and performed extensive ablation studies, comparison of retrieval performance, and analysis of potential uses of the phrase embedding search in few-shot NER scenarios.
We hope that our study will provide practical help in several data-poor domains and valuable insights into entity retrieval and weakly supervised NER.

\section*{Limitations}
Inappropriate initial user questions can negatively affect NER performance. 
If they are not proper, the QA model returns incorrect phrases, and the phrase embedding queries generated from them will also be erroneous. 
The absence of a component for controlling this error cascade in our framework should be addressed in future studies.

In addition, our method is dependent on the phrase encoder of DensePhrases.
Because the phrase encoder is a general-purpose model trained on Wikipedia-based datasets, its capability may be limited for domain-specific entities.
In few-shot NER, the phrase encoder can be sensitive to the quality of given example sentences.
Future studies should thoroughly analyze the effect of the phrase encoder's performance on the resulting NER datasets and NER performance.

\section*{Acknowledgements}
We thank Gangwoo Kim, Miyoung Ko, Donghee Choi, and Jinhyuk Lee for their helpful feedback for the helpful feedback.
This research was supported by (1) National Research Foundation of Korea (NRF-2023R1A2C3004176), (2) the MSIT (Ministry of Science and ICT), Korea, under the ICT Creative Consilience program (IITP-2023-2020-0-01819) supervised by the IITP (Institute for Information \& communications Technology Planning \& Evaluation), and (3) a grant of the Korea Health Technology R\&D Project through the Korea Health Industry Development Institute (KHIDI), funded by the Ministry of Health \& Welfare, Republic of Korea (grant number: HR20C0021(3)).

\bibliography{anthology,custom}
\bibliographystyle{acl_natbib}

\clearpage
\appendix
\setcounter{table}{0}
\setcounter{figure}{0}
\renewcommand\thetable{\Alph{section}.\arabic{table}}
\renewcommand\thefigure{\Alph{section}.\arabic{figure}}

\section{Implementation Details}
\label{appendix:setups}

\paragraph{Input questions.}

We used the same sets of input questions and the same number of sentences for each question as those used in the previous study~\cite{kim2021simple}, which are listed in \Cref{tab:gener_config}.
It should be noted that (1) multiple questions for a single entity type were used because entity types in benchmark datasets are often defined in a coarse-grained way (i.e., they include several sub-categories), and using specific and concrete questions for each sub-category is more effective in covering entities in the benchmark as a whole.
For instance, using three questions, ``Which sports team?'', ``Which company?'', and ``Which institution?'', is better for covering the \textit{organization} type than a single question ``Which organization?''.
In addition, (2) different questions were used for different benchmarks, even though the entity types had the same category names, because the sub-categories were different due to domain and corpus differences between the benchmarks.

\paragraph{Computational environment.}

We ran \ours~and trained all NER models on Intel(R) Xeon(R) Silver 4210R CPU @ 2.40GHz and a single 24GB GPU (GeForce RTX 3090).
When retrieving a huge amount of phrases (e.g., $k_l$ is greater than 100k), we disabled the ``cuda'' option and run the model on the CPU.

\paragraph{Implementation.}

We used the official codes provided by previous studies for the implementation of \bond,\footnote{\url{https://github.com/cliang1453/BOND}} \roster,\footnote{\url{https://github.com/yumeng5/RoSTER}} and \gener.\footnote{\url{https://github.com/dmis-lab/GeNER}}
We used GeNER's repository for the standard models.
We did not implement the few-shot models but used the scores provided by \citet{huang-etal-2021-shot}, \citet{jia-etal-2022-question}, and \citet{kim2021simple}. 
We implemented our phrase embedding search and HighGEN by modifying the code base of GeNER.
We will release our code after the paper is accepted.

\begin{table*}[t!]
\centering
\footnotesize
\begin{adjustbox}{max width = \textwidth}
\begin{tabular}{llccc}
\toprule
\textbf{Dataset} & \textbf{Entity Types (Query Terms)} & \boldsymbol{$k_l$} & \boldsymbol{$k^{\prime}_l$} & \boldsymbol{$|\hat{\mathbf{X}}|$} \\
\midrule
\multirow{3}{*}{CoNLL-2003} & \multirow{1}{*}{\textbf{person}} (athlete, politician, actor) /  & \multirow{3}{*}{5k} & \multirow{3}{*}{30k} & \multirow{3}{*}{45k} \\
& \multirow{1}{*}{\textbf{location}} (country, city, state in the USA) / &  &   \\
& \multirow{1}{*}{\textbf{organization}} (sports team, company, institution) &  &   \\
\midrule
\multirow{3}{*}{Wikigold} & \multirow{1}{*}{\textbf{person}} (athlete, politician, actor, director, musician) / & \multirow{3}{*}{4k} & \multirow{3}{*}{30k} & \multirow{3}{*}{60k}  \\
& \multirow{1}{*}{\textbf{location}} (country, city, state in the USA, road, island) / &  \\
& \multirow{1}{*}{\textbf{organization}} (sports team, company, institution, association, band)   \\
\midrule
\multirow{7}{*}{WNUT-16} & \multirow{1}{*}{\textbf{person}} (athlete, politician, actor, author) /  & \multirow{7}{*}{1k} & \multirow{7}{*}{30k} & \multirow{7}{*}{29k} \\
& \multirow{1}{*}{\textbf{location}} (country, city, state in the USA) /  \\
& \multirow{1}{*}{\textbf{product}} (mobile app, software, operating system, car, smart phone) / &  \\
& \multirow{1}{*}{\textbf{facility}} (facility, cafe, restaurant, college, music venue, sports facility) /  \\
& \multirow{1}{*}{\textbf{company}} (company, technology company, news agency, magazine) / \\
& \multirow{1}{*}{\textbf{sports team}}  (sports team) / \multirow{1}{*}{\textbf{TV show}} (TV show) / \multirow{1}{*}{\textbf{movie}} (movie) / &  \\
& \multirow{1}{*}{\textbf{music artist}} (band, rapper, musician, singer)  \\
\midrule
\multirow{1}{*}{NCBI-disease} & \multirow{1}{*}{\textbf{disease}} (disease)  & 35k & 30k & 35k  \\ 
\midrule
\multirow{1}{*}{BC5CDR} & \multirow{1}{*}{\textbf{disease}} (disease) / \multirow{1}{*}{\textbf{chemical}} (chemical compound, drug)  & 15k & 30k & 45k  \\
\bottomrule
\end{tabular}
\end{adjustbox}
\caption{
Questions and hyperparameters used for NER benchmarks.
Each question is formulated as ``\textit{Which} \typetoken\textit{?}'' and used for the retrieval.
${k_l}$ and ${k^{\prime}_l}$ indicate the number of the top phrases/sentences retrieved from the natural language search and the phrase embedding search for each question, respectively.
${|\hat{\mathbf{X}}|}$ represents the dataset size (i.e., number of training sentences), which is calculated by multiplying the number of questions by $k_l$.
}
\label{tab:gener_config}
\end{table*}

\paragraph{Hyperparameters.}

\begin{itemize}
    \item \textbf{Standard}: 
    Standard models are vulnerable to over-fitting when trained on synthetic data by GeNER or HighGEN. 
    Therefore, we trained \roberta~and \biobert-based models for only one epoch with a batch size of 32 and a learning rate of 1e-5.
    When using full dictionaries, we trained models for ten epochs for CoNLL-2003 and the biomedical domain datasets, and 20 epochs for the other small datasets (Wikigold and WNUT-16).
    \item \textbf{\bond}: We initially trained the teacher model for one epoch and also self-trained the model for additional one epoch.
    For the other hyperparameters, we used the ones suggested by the authors.
    \item \textbf{RoSTER}: We referred to the official repository to select hyperparameters.
    We used the default hyperparameters suggested by the authors, except for noise training epochs and self-training epochs that were set to 1.
    In addition, when training models on biomedical domain datasets by \ours, we used a threshold value of 0.1 in the noisy label removal step.
\end{itemize}

\section{Dataset Statistics}
\label{appendix:statistics}

\begin{table*}[t]
\centering
\footnotesize

\begin{adjustbox}{max width = 0.99\textwidth}

\begin{tabular}{llcccccc}
\toprule
\multirow{2}{*}{\begin{tabular}[c]{@{}l@{}}\textbf{Domain} \textbf{(Corpus)}\end{tabular}} & \multirow{2}{*}{\textbf{Dataset (\# Types)}} & \multicolumn{2}{c}{\textbf{Training}} & \multicolumn{2}{c}{\textbf{Validation}} & \multicolumn{2}{c}{\textbf{Test}} \\
 & & \textbf{\# Sents} & \textbf{\# Labels} & \textbf{\# Sents} & \textbf{\# Labels} & \textbf{\# Sents} & \textbf{\# Labels} \\
\midrule
News (Reuters) & CoNLL-2003 (3)  & 14,987 & 20,061 & 3,469 & 5,022  & 3,685 & 4,947  \\
\cmidrule{1-8}
Wikipedia & Wikigold (3) & 1,142 & 1,842 & 280 & 523 & 274 & 484  \\
\cmidrule{1-8}
Twitter & WNUT-16 (9) & 2,394 & 1,271 & 1,000 & 529 & 3,850 & 2,889  \\
\midrule
\multirow{3}{*}{\begin{tabular}[c]{@{}l@{}}Biomedicine\\(PubMed)\end{tabular}} & NCBI-disease (1) & 5,432 & 5,134 & 923 & 787 & 942 & 960 \\
\cmidrule{2-8}
& BC5CDR (2) & 4,582 & 9,387 & 4,602 & 9,596 & 4,812 & 9,809  \\
\bottomrule
\end{tabular}

\end{adjustbox}

\caption{
Statistics of NER benchmark datasets.
\#~Types: number of entity types. 
\# Sents: number of sentences.
\# Labels: number of entity-level human annotations.
}
\label{tab:statistics}
\end{table*}

\Cref{tab:statistics}~lists the statistics of the five benchmark datasets.

\section{Few-shot Models}
\label{appendix:fewshot}

\paragraph{Supervised\textmd{:}} 
A standard model (described in \Cref{subsec:ner_models}) is trained directly on few-shot examples using a token-level cross-entropy loss.

\paragraph{Noisy supervised pre-training (NSP)~\textmd{\cite{huang-etal-2021-shot}:}}

The model is initially trained on a large-scale weakly-labeled corpus, called WiNER~\cite{ghaddar-langlais-2017-winer}, which consists of Wikipedia documents with weak labels generated using the anchor links and coreference resolution.
Subsequently, the model is fine-tuned on few-shot examples.

\paragraph{Self-training\textmd{~\cite{huang-etal-2021-shot}:}} 

This model is trained using a current semi-supervised learning method~\cite{xie2020self}.
Specifically, the model is initially trained using few-shot examples and fine-tuned by self-training on unlabeled training sentences.
Note that the detailed algorithm can be different from the self-training methods used in \bond~and \roster; therefore, please refer to the papers for details.

\paragraph{\quip~\textmd{\cite{jia-etal-2022-question}:}} 
\quip~was used as the state-of-the-art few-shot model in our experiment.
The model is pre-trained with approximately 80 million question-answer pairs that are automatically generated by the BART-large model~\cite{lewis-etal-2020-bart}, enabling the model to generate high-quality phrase representations, and therefore, achieve strong performance in several few-shot downstream tasks such as NER and QA.
After pre-training, the prediction layer of \quip~is initialized with the embeddings of \textit{question prompts}, which has shown to be more effective in few-shot experiments than random initialization.
For instance, `\textit{Who is a person?}'' was used for the \textit{person} type and ``\textit{What is a location?}'' was used for the \textit{location} type.
We used the same question prompts as those used in the study of~\citet{jia-etal-2022-question} for CoNLL-2003, and those used in the study of~\citet{kim2021simple} for BC5CDR. 

\section{Retrieved Entities}
\label{appendix:retrieved_entities}

\begin{table*}[t!]
\centering
\footnotesize
\begin{adjustbox}{max width = 0.99\textwidth}

\begin{tabular}{ccccc}
\toprule
\multicolumn{4}{c}{\textbf{Natural Language Search}}\\
\midrule
\multicolumn{1}{c}{\textbf{Politician}} & \textbf{Company} & \textbf{Disease} & \textbf{Drug} \\
\cmidrule(lr){1-1} \cmidrule(lr){2-2} \cmidrule(lr){3-3} \cmidrule(lr){4-4}
Ed Miliband & Foxconn &		Leprosy & morphine \\
David Cameron & Boeing &		cirrhosis & opium \\
David Cameron & Plessey &		leprosy & alcohol \\
David Cameron & Marconi &		polio & heroin \\
David Cameron & Sony Corporation &		leprosy & morphine \\
Nick Clegg & Packard Bell &		syphilis & chlorpromazine \\
David Cameron & Airbus &		typhus – & Copaxone \\
David Cameron & Olympus &		Cholera & aspirin \\
David Cameron & Airbus &		syphilis & heroin \\
Douglas Hurd & Airbus &		tuberculosis & Vioxx \\
Ted Heath & Nokia &		typhus & heroin \\
David Cameron & Paramount &		leprosy & imipramine \\
David Cameron & Seagate &		tuberculosis & cocaine \\
Gordon Brown & Cisco &		Leprosy & Thalidomide \\
Gordon Brown & Cisco &		Leprosy & LSD \\
Margaret Thatcher & News Corporation &		syphilis & cocaine \\
Jeremy Corbyn & Nokia &		leprosy & Cisplatin \\
Harold Wilson & Mattel &		polio & penicillin \\
David Cameron & Seagate &		typhus & cannabis \\
David Cameron & Airbus Group &		Measles & Opioids \\
\midrule
\multicolumn{4}{c}{\textbf{Phrase Embedding Search}}\\
\midrule
\multicolumn{1}{c}{\textbf{Politician}} & \textbf{Company} & \textbf{Disease} & \textbf{Drug} \\
\cmidrule(lr){1-1} \cmidrule(lr){2-2} \cmidrule(lr){3-3} \cmidrule(lr){4-4}
David Anthony Laws & Unicer Unicer &  Leprosy & Adrafinil \\
Stefan Löfven & Boeing &		Leptospirosis & Nitrous oxide \\
Michael Ignatieff & Diesel &		hereditary rheumatic syndrome & ivermectin \\
Tony Benn & Arctic &		chronic fatigue syndrome & Pentothal \\
John Major & Monster &		Mal de Débarquement syndrome & Camptothecin \\
Sir Oswald Mosley & Samsung &		seasickness & Glybera \\
George Galloway & Gateway 2000 &		Guillain Barre Syndrome & Trimecaine \\
Arthur Gordon Lishman & Airbus &		Leptospirosis & Gerovital H3 \\
William Hague & Fiat &		Smallpox & Elaterin \\
Sarah Louise Teather & Fiat &		Crohn's disease & Prozac \\
Robert Owen Biggs Wilson & American DeForest &		Achromatopsia & Methamphetamine \\
Helle Thorning-Schmidt & TNT &		Leprosy & metronidazole \\
Philip Andrew Davies & Tenneco Automotive &		Haff disease & Desvenlafaxine \\
Vince Gair & AgustaWestland &		rhabdomyolysis & 4-Fluoroamphetamine \\
Paul William Barry Marsden & Anshe Chung Studios &		Möbius syndrome & ephedra \\
Jeremy William Bray & Raytheon Systems Ltd &		Hansen's Disease & ephedrine \\
Michael Howard & Microsoft &		Lady Windermere syndrome & Alseroxylon \\
Bruce Hawker & Airbus &		McCune–Albright syndrome & Benzydamine \\
Andrew David Smith & Diesel &		Grover's disease & Diclofenamide \\
Peter David Shore & Microsoft &		Lipodermatosclerosis & Cefdinir \\
\bottomrule
\end{tabular}
\end{adjustbox}
\caption{
Top 20 phrases retrieved by the natural language search and phrase embedding search for the four entity types: politician, company, disease, and drug.
}
\label{tab:retrieved_entities}
\end{table*}

\Cref{tab:retrieved_entities} shows the top 20 phrases retrieved by the natural language search and phrase embedding search for the four entity types of politician, company, disease, and drug. 
The phrases from both search methods are generally accurate except for some noisy ones, but the phrase embedding search outperformed the natural language search in terms of the diversity of the retrieved phrases.

\end{document}